\documentclass[3p]{elsarticle}

\usepackage{lineno}
\modulolinenumbers[5]

\usepackage{hyperref}

\usepackage{multirow}
\usepackage{graphicx}

\usepackage[inline]{enumitem}

\journal{Information Sciences}

\makeatletter
\newcommand*\@dblLabelI {}
\newcommand*\@dblLabelII {}
\newcommand*\@dblequationAux {}

\def\@dblequationAux #1,#2,%
    {\def\@dblLabelI{\label{#1}}\def\@dblLabelII{\label{#2}}}

\newcommand*{\doubleequation}[3][]{%
    \par\vskip\abovedisplayskip\noindent
    \if\relax\detokenize{#1}\relax
       \let\@dblLabelI\@empty
       \let\@dblLabelII\@empty
    \else % we assume here that the optional argument
       \@dblequationAux #1,%
    \fi
    \makebox[0.5\linewidth]{%
     \hspace{\stretch1}%
     \makebox[0pt]{$\displaystyle #2$}%
     \hspace{\stretch1}%
    }%
    \makebox[0.5\linewidth]{%
     \hspace{\stretch1}%
     \makebox[0pt]{$\displaystyle #3$}%
     \hspace{\stretch2}%
    }%
    \makebox[0em][r]{(%
  \refstepcounter{equation}\theequation\@dblLabelI, 
  \refstepcounter{equation}\theequation\@dblLabelII)}%
  \par\vskip\belowdisplayskip
}
\makeatother

\makeatletter
\newcommand*\@trplLabelI {}
\newcommand*\@trplLabelII {}
\newcommand*\@trplLabelIII {}
\newcommand*\@trplequationAux {}

\def\@trplequationAux #1,#2,#3,%
    {\def\@trplLabelI{\label{#1}}\def\@trplLabelII{\label{#2}}\def\@trplLabelIII{\label{#3}}}

\newcommand*{\tripleequation}[4][]{%
    \par\vskip\abovedisplayskip\noindent
    \if\relax\detokenize{#1}\relax
       \let\@trplLabelI\@empty
       \let\@trplLabelII\@empty
       \let\@trplLabelIII\@empty
    \else % we assume here that the optional argument
       \@trplequationAux #1,%
    \fi
    \makebox[0.33\linewidth]{%
     \hspace{\stretch1}%
     \makebox[0pt]{$\displaystyle #2$}%
     \hspace{\stretch1}%
    }%
    \makebox[0.34\linewidth]{%
     \hspace{\stretch1}%
     \makebox[0pt]{$\displaystyle #3$}%
     \hspace{\stretch2}%
    }%
    \makebox[0.33\linewidth]{%
     \hspace{\stretch1}%
     \makebox[0pt]{$\displaystyle #4$}%
     \hspace{\stretch4}%
    }%
    \makebox[0em][r]{(%
  \refstepcounter{equation}\theequation\@trplLabelI, 
  \refstepcounter{equation}\theequation\@trplLabelII,
  \refstepcounter{equation}\theequation\@trplLabelIII)}%
  \par\vskip\belowdisplayskip
}
\makeatother

\begin{document}

\begin{frontmatter}

%\title{Elsevier \LaTeX\ template\tnoteref{mytitlenote}}
\title{An Analysis of Hierarchical Text Classification Using Word Embeddings\tnoteref{mytitlenote}}
\tnotetext[mytitlenote]{\copyright 2018. This manuscript version is made available under the CC-BY-NC-ND 4.0 license (\url{https://creativecommons.org/licenses/by-nc-nd/4.0/}). Formal publication available at
\url{https://doi.org/10.1016/j.ins.2018.09.001}.}





\author[Unisinos]{Roger Alan Stein\corref{mycorrespondingauthor}}
%\ead{rogerstein@gmail.com}
\author[Unisinos]{Patr\'{i}cia A. Jaques}
%\ead{pjaques@unisinos.br}
\author[AIE]{Jo\~{a}o Francisco Valiati\corref{mycorrespondingauthor}}
%\ead{joao.valiati@ai-engineers.com}
\address[Unisinos]{Programa de P\'os-Gradua\c{c}\~ao em Computa\c{c}\~ao Aplicada---PPGCA\\ Universidade do Vale do Rio dos Sinos---UNISINOS \\ Av. Unisinos, 950, S\~ao Leopoldo, RS, Brazil}
\address[AIE]{Artificial Intelligence Engineers---AIE\\ Rua Vieira de Castro, 262, Porto Alegre, RS, Brazil}
\cortext[mycorrespondingauthor]{Corresponding authors.}

\begin{abstract}
Efficient distributed numerical word representation models (word embeddings) combined with modern machine learning algorithms have recently yielded considerable improvement on automatic document classification tasks. However, the effectiveness of such techniques has not been assessed for the hierarchical text classification (HTC) yet.
This study investigates application of those models and algorithms on this specific problem by means of experimentation and analysis. We trained classification models with prominent machine learning algorithm implementations---fastText, XGBoost, SVM, and Keras' CNN---and noticeable word embeddings generation methods---GloVe, word2vec, and fastText---with publicly available data and evaluated them with measures specifically appropriate for the hierarchical context. 
FastText achieved an \textsc{lca}F\textsubscript{1} of 0.893 on a single-labeled version of the RCV1 dataset. An analysis indicates that using word embeddings and its flavors is a very promising approach for HTC. 
\end{abstract}

\begin{keyword}
Hierarchical Text Classification
\sep Word Embeddings
\sep Gradient Tree Boosting
\sep fastText
\sep Support Vector Machines
%\texttt{elsarticle.cls}\sep \LaTeX\sep Elsevier \sep template
%\MSC[2010] 00-01\sep  99-00
\end{keyword}

\end{frontmatter}

\section{Introduction}

Electronic text processing systems are ubiquitous nowadays---from instant messaging applications in smartphones to virtual repositories with millions of documents---and have created some considerable challenges to address users new information needs. One of such endeavors is classifying automatically some of this textual data so that information system users can more easily retrieve, extract, and manipulate information to recognize patterns and generate knowledge. Organizing electronic documents into categories has become of increasing interest for many people and organizations \citep{Koller1997Hierarchically, manning2008introduction}.
%provost2013data, ingersoll2013taming}
Text classification (TC)---a.k.a. text categorization, topic classification---is the field that studies solutions for this problem, and uses a combination of knowledge areas such as Information Retrieval, Artificial Intelligence, Natural Language Processing (NLP), Data Mining, Machine Learning, and Statistics. This is usually regarded as a supervised machine learning problem, where a model can be trained from several examples and then used to classify a previously unseen piece of text \citep{sebastiani2002machine, han2011datamining}. 

TC tasks usually have two or a just few classes, for example, automatic email categorization, spam detection, customer request routing, etc. Classification tasks with a high number of possible target classes are studied as a further extension of the TC problem because they present some particular issues, which demand specific addressing or solutions. Many important real-world classification problems consist of a very large number of often very similar categories that are organized into a class hierarchy or taxonomy \citep{manning2008introduction, silla2011survey}. This is where the hierarchical classification (HC) arises: it is a particular type of structured classification problem, where the output of the classification algorithm must correspond to one or more nodes of a taxonomic hierarchy \citep{silla2011survey}.

When applied to textual data, HC then obviously becomes hierarchical text classification (HTC). To illustrate with a real world analogy, HTC is similar to the task of the librarian who needs to find the right shelf for a book from its content. Some examples of large hierarchical text repositories are web directories (e.g.~Best of the Web\footnote{\url{https://botw.org/}}, DMOZ
\footnote{\url{http://dmoz-odp.org/}}
%\footnote{DMOZ---product of the Open Directory Project---was a web directory that used human editors to organize websites. It was closed on March 14, 2017 \citep{sullivan2017ripdmoz}.}
, Wikipedia topic classifications\footnote{\url{https://en.wikipedia.org/wiki/Category:Main\_topic\_classifications}}), library and patent classification schemes (e.g.~Library of Congress Classification\footnote{\url{https://www.loc.gov/aba/cataloging/classification/}}, United States Patent Classification\footnote{\url{https://www.uspto.gov/web/patents/classification/selectnumwithtitle.htm}}), or the classification schemes used in medical applications (e.g.~Medical Subject Headings (MeSH)\footnote{\url{https://meshb.nlm.nih.gov/treeView}}). Many organizations can benefit from automatically classifying documents. For example, law firms can easily place and locate relevant cases \citep{thompson2001automatic},
%, schweighofer2001automatic, governatori2009exploiting}
IT service providers can identify customer needs from incident tickets \citep{zeng2014hierarchical}
%eckstein2016towards}
, medical organizations can categorize reference articles \citep{tsatsaronis2015overview}.
%Trieschnigg2009MeSHUp, peng2016deepmesh}
Some of these examples already take advantage of hierarchical classification structures. Therefore, improvements within the HTC area can have a wide and considerable impact on many applications and areas of knowledge.  

Investigation towards efficient methods to build classification models is of fundamental importance in this context. If a model cannot take advantage of all the training data available or cannot be inducted within a reasonable time, it may not offer an acceptable effectiveness, which in turn may not suit the application needs. The HTC problem poses some particular challenges, and while many classification algorithms are likely to work well in problems with only two or a small number of well-separated categories, accurate classification over large sets of closely related classes is inherently difficult \citep{manning2008introduction}. To address that, some research has been applied onto strategies that exploit the hierarchical structure in problems with a large number of categories. While some results suggest this approach shows some gain over working without using the taxonomy \citep{manning2008introduction, Dumais:2000:HCW:345508.345593} and is overall better than the flat classification approach \citep{silla2011survey}, some conflicting HTC competition results still keep the question open whether hierarchical strategies really outperform flat ones \citep{tsatsaronis2015overview, PartalasKBAPGAA15}. This is therefore a topic that still requires further examination to reach a consensus, as only recently evaluation measures for HTC problems have been better comprehended \citep{kosmopoulos2015evaluation}.
%costa2007review}

Moreover, in the recent years, some breakthroughs have been achieved in the machine learning and NLP fields, which have been improving the effectiveness of many TC systems. Such progress include two main topics: (1) efficient text representation in vector space models such as word embeddings \citep{mikolov2013efficient,pennington2014glove} and (2) efficient classification algorithms implementations, e.g.~softmax-based linear classifiers \citep{Joulin2016fasttext}, scalable tree boosting systems \citep{chen2016xgboost}, and neural network variations \citep{lecun2015deep}. However, to the best of our knowledge, and despite the close relationship between TC and HTC, the impact of those recent advancements have not been fully explored with regards to HTC yet. 

The present work investigated whether and how some techniques that have recently shown to improve the results of TC tasks can be extended to have a positive impact on the HTC problem through empirical experimentation and analysis. More specifically, we have attempted to at least partially address the following main questions:
\begin{itemize}
\item How do recently developed text representation methods---GloVe, word2vec, and fastText---and efficient classification algorithms implementations---fastText, XGBoost, and Keras' CNN---that have recently boosted the flat text classification results improve the effectiveness of HTC?
\item What are the classification models effectiveness difference when comparing traditional classification measures---e.g. flat F\textsubscript{1}---against measures created specifically for hierarchical classification---e.g. hF\textsubscript{1} and \textsc{lca}F\textsubscript{1}?
\end{itemize}

The following three sections provide descriptions of formal HTC definitions~(section~\ref{sec:HTC}), text representation schemes~(section~\ref{sec:LitRev:Text Representation}), and classification algorithms~(section~\ref{sec:LitRev:ClassificationModels}) that we will use for experimentation. Section~\ref{sec:Related Works} reviews relevant advancements within the HTC research, and the impact of recent techniques onto similar classification tasks. Section \ref{sec:Experiments and Analysis} provides a detailed description of the experimental investigation along with its results and an analysis. Finally, section~\ref{sec:Conclusion} summarizes our findings and conclusions.

\section{Hierarchical Text Classification}\label{sec:HTC}

While binary classification is the more general form of TC \citep{sebastiani2002machine}, the current industry needs extend far beyond this fundamental task, which is already challenging in its own way depending on the domain. Some TC tasks can have multiple classes, which can appear in different scenarios. If the classification problem allows for classes that are not mutually exclusive, i.e.~if a text piece can belong to one, more than one, or no class at all, it is called an \textit{any-of}, \textit{multi-value}, or \textit{\textbf{multi{-}label}} classification; on the other hand, if the classes are mutually exclusive, i.e.~each document belongs to exactly one class, it as then called an \textit{one-of}, \textit{single-label}, \textit{multinomial}, \textit{polytomous}, or \textit{\textbf{multi-class}} classification \citep{manning2008introduction}. Throughout the present work, the terms in bold will be preferred. 

If a multi-class task has a large sets of categories, a hierarchical structure is usually present, and taking advantage of it during the learning and prediction processes defines what hierarchical classification is about \citep{silla2011survey}. \citet{Koller1997Hierarchically} were some of the first researchers to notice that the classification schemes that existed at the time ignored the hierarchical structure and were often inadequate in cases where there is a large number of classes and attributes to cope with. This coincides with the emergent popularization of Internet directories such as Yahoo!\footnote{Yahoo! (www.yahoo.com) was created as a directory of websites organized in a hierarchy in 1994. %\citep{Baeza-Yates2011modern}
}, which used to categorize the contents of the World Wide Web. In their proposed approach, they decompose the classification task into a set of simpler problems, and solve each one of them by focusing on a different set of features at each node. 

As hierarchies were becoming ever more popular for the organization of text documents, researchers from the Institute of Informatics and Telecommunications - NCSR Demokritos in Athens, Greece and from the Laboratoire d'Informatique de Grenoble, France organized the Large Scale HTC (LSHTC) Challenge. LSHTC became a series of competitions to assess the effectiveness of classification systems in large-scale classification in a large number of classes, which occurred in four occasions (2009, 2011, 2012, and 2014), and set some benchmarks for the task \citep{PartalasKBAPGAA15}.

\subsection{Problem Criteria and Solution Strategies}\label{subsec:problem_criteria_solution_strategies}

Different HC tasks may have different characteristics that affect how the problem is addressed, such as 
\begin{enumerate*}[label=(\arabic*),series=probcrit1]
\item the hierarchy type\label{item:hiertype}, 
\item the required objective\label{item:hierobj}, 
and \item the way the system uses the hierarchy\label{item:hieruse}. 
\end{enumerate*}
As to \ref{item:hiertype} their type, hierarchical structures are typically trees or direct acyclic graphs---the main difference is that a node can have more than one parent node in the latter. The \ref{item:hierobj} task objective determines whether the classifier must always choose a leaf node---mandatory leaf node prediction (MLNP)---or can choose any node in any level---non-mandatory leaf node prediction (NMLNP) \citep{silla2011survey}. 

The most diverse characteristic relates to \ref{item:hieruse} how a classification system takes advantage of the hierarchy. Many approaches have been proposed to exploit the hierarchical structure of the target categories during the classification processes, and \citet{silla2011survey} summarized them into three main clusters, as follows:
\begin{enumerate}[label=3.\alph*, align=left, labelindent=\parindent, leftmargin=*]
\item{flat}\label{item:flat}: ignores the hierarchy by ``flattening'' it to the leaf nodes level and works any usual multi-class classification algorithm during training and testing phases, 
\item{global}\label{item:global} (a.k.a. big-bang approach): trains a single classifier while taking the hierarchy into account and may use a top-down strategy at the testing phase
\item local approaches\label{item:local}: sometimes incorrectly referred as ``top-down'' approach, uses the hierarchy structure to build classifiers using local information, i.e.~only the data that belongs to a particular node is considered to learn one or many classification models per each node. \citet{silla2011survey} subdivide the local classification approach further into three subgroups depending on the way local information is used at the training phase:
\begin{enumerate}[label=3.c.\roman*, align=left, leftmargin=*]
\item local classifier per node (LCN) trains a binary classifier for each child node
\item local classifier per parent node (LCPN) trains a multi-class classifier for each parent node
\item local classifier per level (LCL) trains a multi-class classifier for the entire hierarchy level
\end{enumerate}
During the test phase, all systems built using this local classification approach use a top-down strategy, i.e.~they predict a class at an uppermost level and then use that information to predict further under the candidates nodes from the previous step only in recursive manner until a leaf node is reached or the blocking criteria for a NMLNP is met. 
\end{enumerate}

%\subsubsection{Evaluation Measures}\label{subsubsec:LitRev:Evaluation}
\subsection{Evaluation Measures}\label{subsec:LitRev:Evaluation}

As hierarchical classification is inherently a multi-class problem, many researchers use traditional multi{-}class evaluation measures such as P (precision, i.e. the percentage of tuples predicted in a given class that actually belong to it), R (recall, i.e. the percentage of tuples correctly classified for a given class), and F\textsubscript{1} measure (a combination of precision and recall in a single measure) \citep{han2011datamining}. As HTC deals with many classes C, a single overall efectiveness value can only be obtained by averaging the mentioned measures, which can be done in two ways, namely, micro-average (average of pooled contingency table) and macro-average (simple average over classes) \citep{manning2008introduction}.

Nevertheless, these measures are actually inappropriate for HTC \textit{as is} because they ignore the parent-child and sibling relationships between categories in a hierarchy, which is intuitively wrong because (1) assigning a tuple to a node near to the correct category is not as bad as assigning it to a distant node, and (2) errors in the upper levels of the hierarchy are worse than those in deeper levels \citep{Sun2001Hierarchical, kiritchenko2006learning, kosmopoulos2015evaluation}. As an attempt to resolve this problem, \citet{Sun2001Hierarchical} proposed two HC measures: a category-similarity based one, which evaluates the effectiveness taking into consideration the feature vectors cosine distance between the correct and the predicted category; and a distance-based one, which assigns effectiveness considering the number of the links between the correct and the predicted category within the hierarchy structure. 

Arguing that these methods are not applicable to directed acyclic graph (DAG) hierarchies nor multi-label tasks, and do not take the node level into consideration to measure the misclassification impact, \citet{kiritchenko2006learning} propose an approach that extends the traditional precision and recall. Instead of considering only the actual and predicted nodes, their measures augment the objects under consideration by considering that each tuple belongs to all ancestors of the class it has been assigned to, except for the root node. The authors call these measures hierarchical precision (hP) and hierarchical recall (hR), which are suitable to calculate a hierarchical F\textsubscript{1} measure (hF\textsubscript{1}) as defined in equations from \ref{eq:hPrecision} to \ref{eq:hF1}. Although they claim having evidences that the new measures are superior to traditional ones, no experimental results have been provided. 

\tripleequation[eq:hPrecision,eq:hRecall,eq:hF1]
{hP = \frac{\sum_{i} |{Anc}_i \cap \hat{Anc}_i|}{\sum_{i} |\hat{Anc}_i|}\mbox{,}}
{hR = \frac{\sum_{i} |{Anc}_i \cap \hat{Anc}_i|}{\sum_{i} |{Anc}_i|}\mbox{,}}
{hF_1 = \frac{2 \cdot hP \cdot hR}{hP + hR}}

where${Anc}_i$ represents all the ancestors of the classes including the true classes and $\hat{Anc}_i$ represents all the ancestors of the classses including the predicted classes.

\citet{kosmopoulos2015evaluation} indicate such hierarchical versions of precision, recall, and F\textsubscript{1} excessively penalize errors in nodes with many ancestors. To address that, they propose a variation, in which they use the lowest common ancestor (LCA) as defined in graph theory---rather than the entire node ancestry as suggested by \citet{kiritchenko2006learning}---to calculate precision (\textsc{lca}P), recall (\textsc{lca}R), and F\textsubscript{1} (\textsc{lca}F\textsubscript{1}) as indicated in equations from \ref{eq:PLCA} to \ref{eq:FLCA}

\tripleequation[eq:PLCA,eq:RLCA,eq:FLCA]
{\textsc{lca}P = \frac{|\hat{Y}_{aug} \cap Y_{aug}|}{\hat{Y}_{aug}}\mbox{,}}
{\textsc{lca}R= \frac{|\hat{Y}_{aug} \cap Y_{aug}|}{Y_{aug}}\mbox{,}}
{\textsc{lca}F_1 = \frac{2 \cdot \textsc{lca}P \cdot \textsc{lca}R}{\textsc{lca}P + \textsc{lca}R}}

where $\hat{Y}_{aug}$ and $Y_{aug}$ are the augmented sets of predicted and true classes. Their \textsc{lca}F\textsubscript{1} measure was studied empirically on datasets used by LSHTC and BioASQ\footnote{BioASQ (\url{http://www.bioasq.org/}) is a challenge in large-scale biomedical semantic indexing and question answering that uses data from PubMed abstracts.} HTC competitions to conclude that ``flat'' measures are indeed not adequate to evaluate HC systems. 

%\subsection{Text Representation}\label{subsec:LitRev:Text Representation}
\section{Text Representation}\label{sec:LitRev:Text Representation}

%Although one can, for example, build a rule-based classifier upon a combination of keywords, most classification systems fundamentally consist of mathematical models that use machine learning-based approaches. Such models are designed to work with examples in the form of \textit{m} tuples containing a fixed number of \textit{n} attributes, ultimately represented as an {\textit{m}-by-\textit{n}} matrix. To comply with this constraint, any raw text that will be submitted to a classification procedure has to be transformed into a compliant representation, known as \textit{vector space model} \citep{sebastiani2002machine, manning2008introduction}. 

In most approaches, TC takes advantage of the techniques developed by the Information Retrieval community to address the document indexing problem in order to build such representation models. Some of these techniques generate a so-called document-term matrix, where each row corresponds to an element of the corpus and each column corresponds to a token from the corpus dictionary \citep{manning2008introduction}, while others represent each document as a vector of an arbitrary size that contains a distribution of representing values, usually called topic models %\citep{blei2012probabilistic}
. A third technique group, more recently developed, computes numerical document representation from distributed word representations previously derived with unsupervised learning methods \citep{mikolov2013distributed, le2014distributed, kusner2015word, balikas2016empirical}. %The following subsections present some of the most relevant text representation methods, which will be either used as benchmark or studied further with this investigation. 

%\subsubsection{Document-term Matrix Approaches}
%\subsection{Document-term Matrix Approaches}

The simplest way to represent text in a numerical format consists of transforming it to a vector where each element corresponds to a unique word and contains a value that indicates its ``weight.'' Such a representation is known in the literature as the \textit{bag of words} (BoW) model \citep{manning2008introduction}. Transforming a document from raw text to BoW usually begins with some data cleansing and homogenization. %In general, techniques such as tokenization (breaking a document into small chunks of text), downcasing (converting the text to lowercase), stemming (shortening a word to its base or root form), and stop words\footnote{Stop words are extremely common words, such as propositions and articles, that have little value for indexing and other text processing tasks, e.g. the, of, in, at, etc. \citep{manning2008introduction}} removal are applied \citep{ingersoll2013taming}.
The next step concerns calculating the weight, which can be done using a wide variety of methods, but usually refers to a composite value derived from the term frequency (TF) and the inverse document frequency (IDF). Both TF and IDF can be calculated in many ways; the most common TF-IDF method attributes the term \textit{t} and document \textit{d} the weight is described as 
%\ref{eq:TF-IDF}
$TF\mbox{-}IDF_{t,d}=tf_{t,d} \cdot \log \frac{N}{df_t}$, where $tf_{t,d}$ is the number of times that term $t$ appears in document $d$, $N$ is the number of documents in the corpus, and $df_t$ is the number of documents that contain term $t$ \citep{feldman2007text}.
%\begin{equation}\label{eq:TF-IDF}
%TF\mbox{-}IDF_{t,d}=tf_{t,d} \cdot \log \frac{N}{df_t}
%\end{equation}
%After calculating the BoW values for all documents in a corpus, each one is then cast into a sparse vector with as many elements as terms in the corpus dictionary to comply with the {\textit{m}-by-\textit{n}} matrix prerequisite. 
Although suitable and efficient for many text mining tasks, BoW models have a very high dimensionality, which poses considerable challenges to most classification algorithms, and ignores the word order completely %\citep{crain2012dimensionality}
. 
%While there are some ways to at least partially tackle the former problem (using feature selection or dimensionality reduction, for example), addressing the latter is impossible as the ordering information loss is irreparable; as a result, classification algorithms will likely predict sentences such ``China attacks France'' and ``France attacks China'' as belonging to the same class, which is obviously wrong if one is trying to classify country belligerence. Furthermore, vector space representations are unable to cope synonymy and polysemy \citep{manning2008introduction}.

%\subsubsection{Distributed Text Representation}
%\subsection{Distributed Text Representation}

Differently of document-term matrix representation, a distributed text representation is a vector space model with an arbitrary number (usually around tens or a few hundreds) of columns that correspond to semantic concepts %\citep{crain2012dimensionality}
. Some of the most popular schemes in this approach %, which is sometimes called \textit{topic modeling}, 
are the latent semantic indexing (LSI) and latent Dirichlet allocation (LDA). LSI consists of a low-rank approximation of the document-term matrix built from it using singular value decomposition (SVD), and can arguably capture some aspects of basic linguistic notions such as synonymy and polysemy;%\citep{deerwester1990indexing}
LDA is a generative probabilistic model in which each document is modeled as a finite mixture of latent topics with Dirichlet distribution \citep{manning2008introduction}.
%\citep{blei2003latent} 

More recent approaches try to compose distributed text representation at document level from word representations in vector space---a.k.a. word vectors or word embeddings---generated with novel methods. Such methods include the continuous bag of words (CBoW) model, the continuous \mbox{skip-gram} model (CSG)---a.k.a. word2vec models\footnote{Many researchers roughly refer to both CBoW and CSG models as \textit{word2vec} models, which is the name of the software implementation provided by \citet{mikolov2013efficient}, which is available at \url{https://code.google.com/p/word2vec/}} \citep{mikolov2013efficient}---and the global vectors model (GloVe) \citep{pennington2014glove}. While word2vec is based on predictive models, GloVe is based on count-based models\citep{Baroni2014}. Word2vec models are trained using a shallow feedforward neural network that aims to predict a word based on the context regardless of its position (CBoW) or predict the words that surround a given single word (CSG) \citep{mikolov2013efficient}. GloVe is a log-bilinear regression model that combines global co-occurrence matrix factorization (somehow similar to LSI) and local context window methods \citep{pennington2014glove}. FastText can be used as a word embedding generator as well; as such, it is similar to the CBoW model with a few particularities, which we described with more details in subsection~\ref{subsec:Linear Classifiers}. Classification algorithms then use the resulting word vectors directly or a combination of them as input to train models and make predictions \citep{huang2014text, kim2014convolutional, lai2015recurrent, balikas2016empirical}. Moreover, some recently proposed classification algorithms incorporate the principles used to compute those word vectors into the classification task itself \citep{le2014distributed, Joulin2016fasttext}.
%tai2015improved
The advantages and disadvantages of the use of these modern text representations remain an open issue.
%\citet{Naili:2017} reported a comparison of word2vec and GloVe in topic segmentation, pointing that depends on the purpose of the application and the idiom, while we bring some pros and cons of each one in the context of HTC (see subsection~\ref{subsec:Exp:Results and Analysis}). 

%/TODO: Waiting for pros and cons... (can be used the citation from Naili et al. (2017))

%\subsection{Classification Models}\label{subsec:LitRev:ClassificationModels}
\section{Classification Models}\label{sec:LitRev:ClassificationModels}

In a broad sense, classification is the process of attributing a label from a predefined set to an object, e.g. classifying an album according to its music genre. In the data mining context though, classification consists of a data analysis in two steps that (1) induces a model from a set of training tuples using statistical and machine learning algorithms that is able to (2) predict which class a previously unseen tuple belongs to \citep{han2011datamining}. As it is a fundamental topic in many areas, classification has been a subject of intensive research over the last decades, which resulted in the proposal of many methods to generate, improve, and evaluate classifiers \citep{manning2008introduction}. The following sub-subsections provide an overview about some of them, namely, linear classifier, gradient tree boosting, and convolutional neural networks (CNN), for future reference in section~\ref{sec:Experiments and Analysis}.

%\subsubsection{Linear classifiers}
\subsection{Linear classifiers}\label{subsec:Linear Classifiers}

A linear classifier assigns class $c$ membership by comparing a linear combination of the features $\vec{w}^{T}\vec{x}$ to a threshold $b$, so that $c$ if $\vec{w}^{T}\vec{x} > b$ and to $\overline{c}$ if $\vec{w}^{T}\vec{x} \leq b$. This definition can be extended to multiple classes by using either a multi-label (\textit{any-of}) or a multi-class (\textit{one-of}) method. In both cases, one builds as many classifiers as classes using a \textit{one-versus-all} strategy. At the test time, a new test tuple is applied to each classifier separately. While all assigned classes are considered for the final result for the multi-label method, in the multi-class only the label with the maximum score $b$ is assigned \cite[p.277--283]{manning2008introduction}. Rocchio\footnote{Rocchio classification model uses centroids to calculate decision boundaries and classify a tuple according to the region it belongs to \cite{manning2008introduction}.}, Na\"ive Bayes\footnote{Na\"ive Bayes is a statistical model based on Bayes' theorem that makes predictions based on the probability that a tuple belongs to a class given its feature values.\citep{han2011datamining}}, and SVM\footnote{Support Vector Machine is a classification model that tries to find the hypothesis that minimizes the classification error based on the structural risk minimization principle by looking for the decision boundary that maximizes the distance between itself and the tuples that belong to either class. \cite{joachims1998text}} are examples of linear classifiers. 

FastText is a model that essentially belongs to this group, but uses a combination of techniques to consider distributed word representation and word order while taking advantage of computationally efficient algorithms \citep{Joulin2016fasttext}. It calculates embeddings in a similar way as the CBoW model does \citep{mikolov2013efficient}, but with the label as the middle word and a bag of \mbox{\textit{n}-gram}s rather than a bag of words, which captures some information about the word order. The algorithm operates in two modes, supervised and unsupervised. In supervised mode, the documents are converted to vectors by averaging the embeddings that correspond to their words and used as the input to train linear classifiers with a hierarchical softmax function %\citep{goodman2001classes}
. On the other hand, in unsupervised mode, fastText simply generates word embeddings for general purposes, then not taking classes into account.

%\subsubsection{Gradient Tree Boosting}
\subsection{Gradient Tree Boosting}

A decision tree is a knowledge representation object that can be visually expressed as upside-down, tree-like graph 
%-- see Figure~\ref{fig:DT}---
in which every internal node designates a possible decision; each branch, a corresponding decision outcome; and a leaf node, the final result (class) of the decision set. If the leaf nodes contain continuous scores rather than discrete classes, it is then called a regression tree. In data mining, decision trees can be used as classification and regression models, and induced from labeled training tuples %through the Hunt's algorithm \citep{%tan2009introduccao, 
%quinlan1986induction}, which constructs the tree 
by recursively partitioning the data into smaller, purer subsets given a certain splitting criteria until either all remaining tuples belong to the same class, there are no remaining splitting attributes, or there are no remaining tuples for a given branch \citep{han2011datamining}.

There are many ways to improve the tree induction algorithm by, for example, using different splitting criteria (gain ratio, information gain, Gini index, $\chi^2$, etc.), pruning too specific branches, or using tree ensembles. Boosting is an ensemble method in which a classifier $M_{i+1}$ is learned by ``paying more attention'' to the training tuples that were previously misclassified by $M_i$ \citep{han2011datamining}. The final classification is done by combining the votes of all $M$ classifiers weighted by each model corresponding accuracy \citep{han2011datamining}. Gradient boosting is a method that creates an ensemble of weak regression trees by iteratively adding a new one that improves the learning objective further through optimization of an arbitrary differentiable loss function 
%\citep{friedman2001greedy}
. A recent implementation of this method called XGBoost\footnote{The system is available as an open source package at \url{https://github.com/dmlc/xgboost}} combines computationally efficient principles---parallelism, sparsity awareness, cached data access---with additional improvement techniques, and has been allowing data scientists to achieve state-of-the-art results on many machine learning challenges \citep{chen2016xgboost}.

%\subsubsection{Neural Networks}\label{subsubsec:LitRev:Neural Networks}
\subsection{Neural Networks}\label{subsec:LitRev:Neural Networks}

A neural network (NN) is a set of units in the form of a DAG, where each unit node processes a function, and each connection has a weight associated with it \citep{han2011datamining, Goodfellow-et-al-2016}
%freeman1991algorithms,
. The most popular NN architecture is the multilayer feed-forward, in which the units are organized in three parts: the input layer, which receives the external data; the hidden layer, which might consist of many levels, indicating the depth of the network; and the output layer, which emits the network's prediction. NN's are most commonly trained by backpropagation, a method that iteratively updates the network connection weights to minimize the prediction errors \citep{han2011datamining}.

Even though the interest on NN was less intense during some decades, it has been a topic of continuous research since its first appearance in the 1940s, and it has been drawing considerable attention due to the recent emergence of deep learning (DL) techniques over the last years \citep{schmidhuber2015deep}. Although having many hidden layers is a common characteristic of DL architectures, their key aspect is actually the fact that they allow for the representations of data with multiple levels of abstraction. This allowed DL to produce extremely promising results for various tasks in natural language understanding, particularly topic classification \citep{lecun2015deep}. Besides the general feed-forward neural network (FNN), a few specialized architectures are already used heavily in industry, including CNN and recurrent neural networks (RNN), which can scale to, for example, high resolution images and long temporal sequences \citep{Goodfellow-et-al-2016}.

CNN is a specialization of FNN that employs convolution---a specialized kind of linear operation---rather then a matrix multiplication with connection weights. Its architecture usually consists of layers with three stages, namely, convolution, detection, and pooling.
%, as shown in Figure~\ref{fig:CNN}
Despite its name, the convolution stage does not necessarily execute the title operation as mathematically defined; it applies a function over the input using a kernel that works as a kind of image filter resulting in a set of linear activations. The detection stage runs a nonlinear activation function over those previous results, usually a rectified linear activation. Finally, the pooling stage replaces the detection output with a summary statistic of nearby outputs, which might be used to make a final prediction or to connect to the next convolution layer \citep{Goodfellow-et-al-2016}.

%====================================================================
\section{Related Works}\label{sec:Related Works}

The problem at hand has been a widely researched topic over the last two decades, and many approaches have been attempted towards the improvement of the classification results. At the same time, recent investigation on problems that bear some similarity  with the HTC, such as binary TC (sentiment analysis, spam detection, etc) have experienced some rapid development with the usage of the representation and classification methods presented in sections~\ref{sec:LitRev:Text Representation} and \ref{sec:LitRev:ClassificationModels}.  

This section aims to present past and current research status regarding HTC and some techniques used in related areas that can have an impact on this field as well, considering the similarity that it holds with other TC problems. Subsection~\ref{subsec:RelatedWorks:HTC} provides an overview about recent HTC research, sections~\ref{subsec:RelatedWorks:TextClasDistTextRep} and \ref{subsec:RelatedWorks:NeuralNets} describe the advancements that other TC problems have seen in recent years, and finally subsection~\ref{subsec:RelatedWorks:Discussion} critically analyzes all those studies and their relation to HTC. 

\subsection{Hierarchical Text Classification Research}\label{subsec:RelatedWorks:HTC}

\citet{Koller1997Hierarchically} proposed probably the first model that took the hierarchical structure of the target categories into consideration to build a classification system. It consisted of a set of Bayesian classifiers, one at each hierarchy node, which would direct a new incoming test tuple that made it through the parent nodes to the proper child node. Before being processed through the classifier, the text was submitted to a Zipf's Law-based\footnote{Zipf's Law is an empirical rule% formulated by \citet{zipf1935psycho} 
that states that the collection frequency $cf_i$ of the \textit{i}th most common term is proportional to $1/i$ \cite[p.82]{manning2008introduction}.} filter and encoded as a boolean vector. Experimental results using the Reuters-22173\footnote{\url{https://archive.ics.uci.edu/ml/datasets/reuters-21578+text+categorization+collection}} showed a significantly higher 
then previous models due to (1) mainly the selection of features and (2) marginally the  hierarchical disposition of the individual classifiers, as long as they are complex ones---i.e.~the benefit was inconclusive while using Na\"ive Bayes model, but substantial with a more elaborated algorithm from the Bayesian family.
%called KBD \citep{sahami1996learning}.

Soon after that, \citet{Dumais:2000:HCW:345508.345593} used SVM to build an HTC model with two levels. The classification is based on threshold, and considers parent and child nodes either in a Boolean decision function (LCN approach) or multiplying them (LCPN approach). In other words, the model would classify a tuple as belonging to a node if the calculated probability for it was higher than a user-specified value. The authors used SVM because it was considered an effective, efficient algorithm for TC, and experimented on a web content data set with 350K records, which was a considerable amount for the time. The results showed a small F\textsubscript{1} improvement over flat models, but still statistically significant. On the other hand, a very recent study suggests that hierarchical SVM results do not considerably differ from the corresponding flat techniques \citep{graovac2017hierarchical}. %\citet{cesa2006hierarchical} tried to combine both Bayesian and SVM models, but the results were unclear about any considerable advantage brought by their approach. 

\citet{ruiz2002hierarchical} considered using an NN to create a model for HTC. Their collection of feedforward neural networks was inspired by a previous work 
%on Hierarchical Mixture of Experts \citep{jordan1994hierarchical}, 
and consisted of a tree-structure composition of expert (linear function) and gating (binary function) networks trained individually. The experiments were executed on an excerpt of 233,455 records with 119 categories from the OHSUMED collection\footnote{\url{http://davis.wpi.edu/xmdv/datasets/ohsumed}}.
%\citep{hersh1994ohsumed}
The data was filtered by stop words, stemmed using Porter's algorithm, underwent feature selection through correlation coefficient $\chi^2$, mutual information, and odds ratio, and then was finally submitted to the model. When comparing the results against a flat model, the hierarchical model results (as measured by an F\textsubscript{1} variation) are better, which indicates that exploiting the hierarchical structure increases effectiveness significantly. Nevertheless, the proposed approach is only equivalent to a Rocchio approach, which was used as benchmark. 

In the realm of boosting methods, \citet{esuli2008boosting} propose TreeBoost.MH, which is a recursive, hierarchical variant of AdaBoost.MH, a then well-known, multi{-}label algorithm that iteratively generates a sequence of weak hypotheses to improve upon it.
%\citep{schapire1998improved}.
The researchers experimented the method on Reuters-21578 (90 classes, $\sim$11K records), the RCV1\footnote{Reuters Corpus Volume 1 \citep{lewis2004rcv1}} (103 classes, $\sim$800K records), and the ICCCFT\footnote{2007 International Challenge on Classifying Clinical Free Text Using Natural Language Processing} (79 classes, $\sim$1K records), and considered the same F\textsubscript{1} function variation as \citet{ruiz2002hierarchical} did for an effectiveness measure. Their conclusion is that the hierarchical AdaBoost.MH variant substantially surpasses the flat counterpart, in particular for highly unbalanced classes. Nonetheless, they mention that their approach is still inferior to SVM models, but make reservations regarding the validity of such a comparison. 

The editions of the LSHTC Challenge brought many diverse approaches into the HTC area. \citet{PartalasKBAPGAA15} report on the dataset construction, evaluation measures, and results obtained by participants of the LSHTC Challenge. The most important dataset (used in 3 of the four editions) consisted of 2.8M records extracted from DBpedia\footnote{\url{http://wiki.dbpedia.org/}} instances distributed among 325K classes. Instead of the original text from the DBpedia instance, each record consisted of a sparse vector with (feature, value) pairs resulting from a BoW processing. The challenge organizers used many evaluation measures, including accuracy, precision, recall, F\textsubscript{1} measure, and some hierarchically-specialized ones introduced over the years by \citet{kosmopoulos2015evaluation}, but not reported in this competition overview. The report summarizes the results by saying that flat classification approaches were competitive with the hierarchical ones, and highlights only a few that seem noteworthy, such as some models built upon \textit{k}-Nearest Neighbor (\textit{k}NN)\footnote{Nearest Neighbors classifiers are labor intensive classification methods that are based on comparing a given test tuple with training tuples that are similar to it \cite[p.423]{han2011datamining}} and Rocchio improvements. The 4th edition winning submission, in particular, consisted of an ensemble of sparse generative models extending Multinomial Na\"ive Bayes that combined document, label, and hierarchy level multinomials with feature pre-processing using variants of TF-IDF and BM25 \citep{PuurulaRB14}. 

\citet{balikas2016empirical} elaborated an empirical study that employed word embeddings as features for large scale TC. The researchers considered three versions with 1K, 5K, and 10K classes of a dataset with 225K records originally produced for the BioASQ competition, in which each record corresponds to the abstract, title, year, and labels of a biomedical article \citep{tsatsaronis2015overview}. Despite using datasets with that high number of classes, these are not considered in a hierarchical fashion, which means the task consists of a flat, multi{-}label classification. The word embeddings were generated using the skip{-}gram model of word2vec based on 10 million PubMed\footnote{\url{https://www.ncbi.nlm.nih.gov/pubmed/}} abstracts plus 2.5M Wikipedia documents in four sizes: 50, 100, 200, and 400 elements. The resulting embeddings of all words in a document abstract were combined using different functions---min, max, average, and a concatenation of these three---to compose a document representation, and the resulting vector was then used as input into an SVM classifier. The best results (as measured by F\textsubscript{1}) are achieved by the concatenation of the outputs of the three composition functions, and are consistently better than the runner up, the average function. Besides the way the word embeddings are combined, the vector size has proportional effect on the classification effectiveness as well. The results, however, do not reach the baseline model, which is TF-IDF based SVM model. Nevertheless, when combining TF-IDF to the concatenated document distributed representations, the results are better than the TF-IDF alone by a small, but statistically significant margin.

\subsection{Text Classification with Distributed Text Representations}\label{subsec:RelatedWorks:TextClasDistTextRep}

While no breakthrough has occurred with the HTC task over the last years, other TC problems on the other hand have benefited from the recent great improvements on text representation using distributed vector space models. Over the recent years, many researchers used methods based on word embeddings to improve the accuracy of classification tasks such as sentiment analysis and topic classification. This section provides some examples from these other TC tasks that are somehow similar to HTC and took advantage from those advancements.

With regards to sentiment analysis, for example, \citet{Maas2011} created a method inspired on probabilistic topic modeling to learn word vectors capturing semantic term{-}document information with the final intend to tackle the sentiment polarization problem. They collected a dataset\footnote{The so-called Large Movie Review Dataset v1.0 has been widely used as a benchmark dataset for binary sentiment TC and is publicly available at \url{http://ai.stanford.edu/\~amaas/data/sentiment/}} with 100,000 movie reviews from the Internet Movie Database (IMDb)---25,000 labeled reviews for the classification task, 25,000 for the classification test, and 50,000 of unlabeled ones as additional data to build the semantic component of their model. The semantic component consisted of 50-dimensional vectors learned using an objective function that maximizes both the semantic similarities and the sentiment label. Their model outperformed other approaches, in particular when concatenated with BoW, when compared results upon the \textit{polarity dataset v2.0}\footnote{A dataset with 2,000 balanced, processed reviews from the IMDb archive publicly available at \url{http://www.cs.cornell.edu/people/pabo/movie-review-data/}}.
%\citep{pang2004sentimental}

\citet{le2014distributed} proposed an unsupervised learning method that calculates vectors with an arbitrary length containing distributed representations of texts with variable length---the so-called \textit{paragraph vectors}, which was highly inspired by the techniques used to learn word vectors introduced by \citet{mikolov2013distributed}. Such paragraph vectors can be used as features for conventional machine learning techniques, so the authors took the data collected by \citet{Maas2011} to calculate paragraph vectors, used them as inputs to a neural network to predict the sentiment, and compared the results against other approaches that used the same dataset. They reported a final result of 7.42\% error rate, which they claim meant a new state-of-the-art result, with a significant relative error rate decrease in comparison to the best previously reported method.

Still on the sentiment analysis topic, however on a slightly different scenario, \citet{tang2014learning} proposed the learning of Sentiment Specific Word Embedding (SSWE) by integrating the sentiment information into the loss function of the model and its application in a supervised learning framework for Twitter sentiment classification task. This is similar to the idea proposed by \citet{Maas2011}, but uses a neural network rather then a probabilistic model. The authors used a partial version of a benchmark dataset used on SemEval\footnote{SemEval (Semantic Evaluation) is an ongoing series of evaluations of computational semantic analysis systems organized by the Association for Computational Linguistics (ACL) \url{https://aclweb.org/aclwiki/SemEval_Portal}} 2013 %\citep{nakov2013semevaltask2}
with 6,251 positive/negative unbalanced records, and found that the SVM classification model built upon their SSWE has an effectiveness (macro-F\textsubscript{1}) comparable with models created from state-of-the-art, manually designed features. Furthermore, they compared their SSWE with three other word embeddings---C\&W\footnote{C\&W is a short that \citet{tang2014learning} used for the method reportedly introduced by \citet{collobert2011natural}, which was not formally named by the authors.}%, but apparently first appeared in \citet{collobert2008unified}
\citep{collobert2011natural}, word2vec \citep{mikolov2013efficient}, and WVSA (Word Vectors for Sentiment Analysis) \citep{Maas2011}---to conclude that the effectiveness of word embeddings that do not directly take advantage of the sentiment information in the text---C\&W and word2vec---are considerably lower than the others. Their study is just the beginning of a clear strategy trend in this topic: 7 out of the 10 top-ranked solutions for the SemEval-2016 Sentiment Analysis in Twitter Task incorporated either general-purpose or task-specific word embeddings in their participating systems \citep{nakov2016semevaltask4}. As an exponent of this trend, \citet{vosoughi2016tweet2vec} created a method to compute distributed representations for short texts using a long short-term memory (LSTM)\footnote{The long short-term memory (LSTM)
%was originally proposed by \citet{hochreiter1997long},
uses ``a memory cell which can maintain its state over time and non-linear gating units which regulate the information flow into and out of the cell'' \citep{greff2016lstm}, and is usually associated with the deep learning algorithms family \citep{lecun2015deep}.} NN at a character-level. As their data source was the microblog Twitter, they adequately named the method \textit{Tweet2Vec}. The model was trained with 3 million records, which consisted of texts with at most 140 characters. To evaluate the quality of the resulting vectors, the authors used them to perform a polarity classification on the dataset provided on SemEval-2015 Task 10 subtask B competition\footnote{\url{http://alt.qcri.org/semeval2015/task10/}}. The experiment consisted on extracting the vector representation from the texts on that dataset using their method and then train a logistic regression classifier on top of them. Their approach has reportedly exceed all others from that competition, and also surpassed \citet{le2014distributed} \textit{paragraph vector}, which was considered the state of the art in that context. 

On the other hand, there are also examples of the usage of word embeddings on more general, multi category TC tasks. \citet{huang2014text} propose a method to learn so called \textit{document embeddings} directly in TC task, that aims to represent a document as a combination of the word embeddings of its words, which is learned using a neural network architecture. The authors use resulting network in two ways during the classification phase: the network itself as a classification model or the weights from one of its last hidden layers as the input for an SVM classifier. They test their methods on two datasets with 9 and 100 categories, and 17,014 and 13,113 training records, respectively---interestingly enough, the dataset with more categories was extracted from the LSHTC Challenge 4th edition, but ignored its hierarchical characteristic and used documents with a single label only. Although the authors report that their proposed architecture achieve better effectiveness on both datasets, the difference is only evident in one of them, and not enough statistical information is provided to support that claim. 

\citet{ma2015distributional} use a Gaussian process approach to model the distribution of word embeddings according to their respective themes. The authors assume that probability of a document given a certain theme is the product of the probabilities of a word vector given the same theme. The classification task then becomes a problem of selecting the most probable Gaussian distribution that a document belongs to. The authors evaluate the model effectiveness with a dataset containing 10,060 training and 2,280 test short texts that belong to 8 unbalanced classes, which was previously used by other researchers. Their results show that the proposed method has a 3.3\% accuracy gain over two other approaches that used (1) classical TF-IDF and (2) topic models estimated using latent Dirichlet allocation (LDA) as representation methods connected to MaxEnt classifiers, and suggest the accuracy increase occurs because ``It is clear that using word embeddings which were trained from universal dataset mitigated the problem of unseen words.'' %Nevertheless, accuracy is not an adequate metric for an unbalanced, multi-class classification \citep{fawcett2006introduction}.

On another approach, \citet{kusner2015word} take advantage of the word embeddings to create a distance metric between text documents. Their proposed metric aims to incorporate the semantic similarity between word pairs---the lowest ``traveling cost'' (Euclidean distance) from a word to another within the word2vec embedding space---into a document distance function. The minimum cumulative cost to move from a document to another---the so-called \textit{Word Mover's Distance} (WMD)---is then used to perform \textit{k}NN document classification on eight real world document classification data sets. The resulting \textit{k}NN classification model using WMD yields unprecedented low classification error rates when compared to other well established methods such as latent semantic indexing (LSI) %\citep{deerwester1990indexing} 
and LDA. %\citep{blei2003latent}

\citet{Joulin2016fasttext} built a classification model called fastText, already presented in section \ref{sec:LitRev:ClassificationModels}. The researchers ran experiments with two different tasks for evaluation, namely sentiment analysis and tag prediction. For sentiment analysis comparison, they used the same eight datasets and evaluation protocol as \citet{zhang2015character}, and found that fastText (using 10 hidden units, trained 5 epochs, with bigram information) accuracy is competitive with complex models, but needed only a fraction of time to process---the faster competitor took 24 minutes to train (some took up to 7 hours), while the worst case for fastText took only 10 seconds. Moreover, the authors claim they can still increase the accuracy by using more \mbox{\textit{n}-gram}s, for example with trigrams. For tag prediction evaluation, they used a single dataset\footnote{YFCC100M dataset -- \url{http://yfcc100m.appspot.com/} %\citep{thomee2016yfcc100m}
} that contains information about images and focused on predicting the image tags according to the their title and caption. 
%When compared to Tagspace model \citep{weston2011wsabie}, 
FastText has a significantly superior effectiveness (as measured by precision-at-1)---because it is not only more accurate, but also uses bigrams---and runs more than an order of magnitude faster to obtain the model. To summarize, the fastText shallow approach seems to obtain effectiveness on par with complex deep learning methods, while being much faster.

\subsection{Neural Networks for Text Classification}\label{subsec:RelatedWorks:NeuralNets} 

Neural network models have been investigated for TC tasks since mid 1990s \citep{schutze1995comparison, wiener1995neural}. Nevertheless, at the same time that \citet{sebastiani2002machine} reports that some NN-based models using logistic regression provide some good results, he observes that they have a relative effectiveness slightly worse than many other models known at the time, e.g.~SVM. This scenario would not change much during the following decade, such is that an evidence of NN models unpopularity in this area is the fact that \citet{manning2008introduction} do not even mention them in their classic textbook. Its resurgence among the TC research community would begin only in the late 2000's \citep{zhang2006multilabel, trappey2006development} %\citep{yu2008latent} 
and maintain a steep increase over the following years, as will be shown in the upcoming paragraphs.

%In a follow-up of the work described in their \citeyear{collobert2008unified}'s paper, 
\citet{collobert2011natural} bring some new, radical ideas to the natural language processing area while deliberately disregarding a large body of linguistic knowledge to propose a neural network and learning algorithm that, contrary to the usual approach, is not task-specific, but can be applied to various tasks. In the authors' point of view, part-of-speech tagging (POS), chunking, named entity recognition (NER), and semantic role labeling (SRL) problems can be roughly seen as assigning labels to words, so they build an architecture capable of capturing feature vectors from words and higher level features through CNN to do that from raw text. Their training model architecture produces local features around each word of a sentence using convolutional layers and combines these features into a global feature vector, which is then fed into a standard layer for classification. They compare the results using this architecture against the state-of-the-art systems for each one of the four traditional NLP tasks just mentioned, and find that the results are behind the benchmark. To improve them, the authors create language models using additional unlabeled data that obtain feature vectors carrying more syntactic and semantic information, and use them as input for the higher level layers of the architecture. This approach not only brings the results a lot closer to those benchmark systems', but demonstrates how important the use of word embeddings learned in an unsupervised way is. Not satisfied, \citet{collobert2011natural} still use some multi-task learning schemes to have even more, better features, and eventually use some common techniques from the NLP literature as a last attempt to surpass the state-of-the-art systems. Their final standalone version of the architecture is a ``fast and efficient `all purpose' NLP tagger'' that exceeds the effectiveness of the benchmark systems in all tasks, except for semantic role labeling, but only with a narrow margin.

%\citet{socher2013recursive} proposed a model called Recursive Neural Tensor Network (RNTN) to compute compositional vector representations for phrases of variable length that are then used as features for sentiment classification. The input data consists of phrases parsed into binary trees with each leaf node corresponding to a word (and its corresponding vector). The model is based upon a standard recursive neural network (RcsNN), which computes parent vectors in a bottom up fashion with different composition functions using the same softmax classifier; the RNTN however differs itself from those by introducing a tensor-based composition function. The authors introduced the Stanford Sentiment Treebank, a fully labeled parse trees phrase dataset, to accurately analyze the model at phrase level. The new model was compared to with other recursive methods, namely standard RcsNN and Matrix-Vector RcsNN (MV-RcsNN), and some commonly used methods---Na\"ive Bayes and SVMs using bag of words and bag of bigrams representations. The authors find that the recursive methods achieve better results than common methods, particularly because the bottom up approach is able to  capture negations at many levels in both positive and negative phrases. Finally, the proposed new method supplanted all others known at the time, and pushed the state-of-the-art accuracy on that dataset to 85.4\%.

Still on sentiment classification studies, \citet{kim2014convolutional} reports on experiments with CNN trained upon distributed text representations. This approach is similar to previously mentioned architecture \citep{collobert2011natural}, but uses pre-trained word embeddings learned with word2vec and proposes a modification to allow for the use of both pre-trained and task-specific vectors by having multiple channels. The experiments included 7 datasets that not only related to sentiment polarity, but also considered subjectivity and question type classification, with number of classes between from 2 and 6, and dataset size ranging from 3.8 to 11.9 thousand records. The experiment results show that the authors' simple CNN with one convolution layer only performs remarkably well despite little tuning of hyperparameters, surpassing the state-of-the-art methods in 4 out of the 7 analyzed tasks/datasets. It is not clear, however, what was the exact standard used to evaluate the effectiveness, nor whether the difference had a statistical significance, which is important as the tests related to 3 of the 4 superior scenarios were executed using a 10-fold cross-validation.
 
\citet{johnson2014effective} also use CNN architecture, but instead of word embeddings, their model works on high-dimensional one-hot encoding vectors, i.e.~each document is seen as sequence of dictionary-sized, ordered vectors with a single true bit each that corresponds to a given word. Their intention with this approach is capturing the word order within the network convolution. For evaluation purposes, the authors executed experiments on sentiment classification---IMDb dataset \citep{Maas2011}---and topic categorization---RCV1 dataset \citep{lewis2004rcv1}, disregarding the hierarchical structure --, and compared the results against SVM-based classifiers. Their CNN
allegedly outperforms the baseline methods as measured by error rate, but this claim lacks of some substantial statistical analysis. On a similar idea, but with a more minimalistic approach, the model proposed by \citet{zhang2015character} accepts a sequence of encoded characters as input to a convolutional network to classify documents. In other words, the researchers deliberately disregard the grouping of letters in words, and transform text at character level into a 1,024 fixed length flow of vectors created with one-hot encoding. Since such kind of model requires very large datasets, in order to perform meaningful experiments, the authors had to build 8 of them, which had from 2 to 14 classes, number of records ranging from 120,000 to 3.6 million, and related to two main task, namely sentiment analysis and topic classification. To compare the models effectiveness, besides training their own new model, the authors also did so with models using (1) a multinomial logistic regression method built upon traditional text representation techniques and (2) a recurrent neural network using pre-trained word2vec word embeddings as input. In conclusion, they found that character-level convolutional networks is a feasible approach for TC, and confirmed that such models work best having large datasets for training. 

\citet{lai2015recurrent} combine recurrent and convolutional NN's to tackle the TC problem. At the model first level, a bi-directional recurrent structure captures the contextual information; at its second level, a max-pooling layer finds the best features to execute the classification. The model input consists of word embeddings that were pre-trained using the word2vec \mbox{skip-gram} method on Wikipedia dumps. For experimentation, the authors used 4 multi-class datasets with different sizes, and compared the model result in each dataset against the state-of-the-art approach for each dataset. Their model performs consistently well in all tested datasets, and even beats the best performing ones in half of the cases by a considerable difference, leading to the confirmation that NN-based approaches can compute an effective semantic text representation, and their conclusion that such approaches can also capture more contextual information of features than traditional methods.

\subsection{Discussion and Considerations}\label{subsec:RelatedWorks:Discussion}

Considering the works mentioned in this section, a few trends seem evident. The first thing to notice with regards to HTC research is that no reference dataset has apparently emerged over these two decades, and despite a few appear more often, there is no widely used standard. This obviously impedes effectively comparing the results of different studies in any way, as using the same data for experimentation is a prerequisite for any analysis with this intention. Although the Reuters' collections seemed to become popular at some point, they was disregarded by the LSHTC Challenge, which probably demanded a larger text collection. Nowadays even LSHTC dataset seems small, and its preprocessed format has actually become an inconvenient for researchers who intent to use distributed text representation. On a second note, no single effectiveness measure has been widely accepted yet as well. This is in part because of the variations within the HTC task itself (single- or multi{-}label), but also in part because it seems it took a long time for the community to evolve to a point when a thorough study about hierarchical classification evaluation could have been done. This fact not only poses a problem to compare the results among different studies, but also suggests that the comparison against flat models is not possible, as the measure for one problem is simply not the same as for the other. In other words, comparing the effectiveness of hierarchical classification against flat classification is not only inadequate, but also inaccurate, as the problem is different in its own nature \citep{kosmopoulos2015evaluation}. In summary, the lack of consensus regarding a reference dataset and evaluation measures has negatively affected the HTC research development.

Many different methods have been applied to HTC, and the most representative ones have been referred, namely Bayesian models \citep{Koller1997Hierarchically, PuurulaRB14}, SVM \citep{Dumais:2000:HCW:345508.345593, graovac2017hierarchical}, NN \citep{ruiz2002hierarchical}, boosting methods \citep{esuli2008boosting}, Rocchio, and \textit{k}NN \citep{PartalasKBAPGAA15}. This list is nonetheless far from exhaustive, as any text classification method (and any classifier in general, by extension), could be virtually used in this context. Nevertheless, neither a consistent effectiveness increase nor a breakthrough seem to have occurred over these two decades in the area. It is interesting to notice how \citet{esuli2008boosting} consider the improvement achieved by their hierarchical model somehow surprising, as they would expect some effectiveness counter effect due to the unrecoverable incorrect classifications that occur in lower hierarchy levels, which is a known side-effect since \citet{Koller1997Hierarchically}.

The fact that recent results still do not show a considerable difference between flat and hierarchical classification \citep{PartalasKBAPGAA15,graovac2017hierarchical} sounds disquieting, to say the least, as one would expect that a specialized system should behave better than a generic one. Of course, the direct comparison does not hold, and should not be made. However, the proximity between flat and hierarchical classification makes it inevitable. Such comparison, on the other hand, makes the fact that HTC researchers seem to have been paying little or no attention to the recent classification effectiveness improvements achieved using advances in other TC tasks also surprising. For example, considering the reports on TC competitions, while \citet{PartalasKBAPGAA15} do not even refer to the usage of word embeddings in LSHTC Challenges, \citet{nakov2016semevaltask4} mention that most of the best systems performing sentiment analysis on Twitter used them in some way. However, after some consideration, it becomes clear that such advanced techniques could not had been applied to the LSHTC Challenge due to the way the dataset has been provided. Apparently, some data cleansing processing was so widely accepted around the competition years that the documents were heavily preprocessed (stemming/lemmatization, stop-word removal), and the idea of BoW was so well established that the organizers decided to deliver the dataset in a sparse vector format where each line corresponded to a document, and contained only token identifiers with its corresponding frequency. Although very popular and quite effective, this format misses some important lexical richness and lacks the word order, which is an overriding factor to detect semantic nuances.

It seems clear that word embeddings and other vector space models improve some TC schemes considerably. In the sentiment analysis task, techniques that either create vector space models in a supervised, polarity-induced manner \citep{Maas2011,socher2013recursive,tang2014learning} or use general-purpose models \citep{kim2014convolutional,le2014distributed}
%li2016tweet
benefit from them. Similar advantage is reported in more general problems \citep{huang2014text,ma2015distributional}, although some healthy skepticism is advisable regarding those reports as the evaluation methods are questionable. Nevertheless, the ideas behind word embeddings are undoubtedly advantageous for TC in many different ways, from calculating a distance metric for \textit{k}-NN classifier \citep{kusner2015word} to transforming a word embedding learner into a classifier itself \citep{Joulin2016fasttext}. \citet{balikas2016empirical} mention they are aware that word embeddings are sometimes used as input for convolutional and recurrent neural network, but as their task concerns a large number of classes, they refrained from using them to avoid computational obstacles such as memory and processing overhead. The workaround they used, i.e.~combining the word embeddings with simple arithmetic functions, yields good results, but still ignores the word order. All in all, despite its promising results, the effect of using word embeddings in HTC remains a great unknown, as no empirical evidence has been reported on it.  

Analogously, it is evident that many TC and NLP tasks have been taking advantage of recent neural network architectures and deep learning improvements. Although \citet{collobert2011natural}
%collobert2008unified
do not work with TC at sentence or document level, the ideas proposed therein seem significantly influential considering that many of the neural network architectures used today for that task had some inspiration taken from them\footnote{Those two papers combined had more than 3,500 citations as counted by Google Scholar by Jan 2017}. Although \citet{collobert2011natural} show that the use of CNN provides competitive results in more than one NLP classification task, the concepts that have been preached by them and others influenced many following researchers who later on started to reconsider NN models and find promising results \citep{tang2014learning, kim2014convolutional}. Their most important contribution to the present investigation is the indication that adequate word embeddings combined with appropriate classification NN's provide promising results. 

On top of that, comparisons using simple feed-forward NN's against other methods have shown that the former are not only competitive, but even outperform the latter in many cases. This has been confirmed time and again with more complex architectures such as recursive NN's \citep{socher2013recursive}, recurrent NN's \citep{zhang2015character}, convolutional NN's \citep{kim2014convolutional}, or a combination of them \citep{lai2015recurrent}. All these works corroborate to the belief that a neural network is the most appropriate architecture to implement a state-of-the-art classification system. Nevertheless, this assumption lacks of empirical evidence when it comes to the hierarchical text context, as no report has been found specifically about it and it is doubtful that simple TC problems are adequate to evaluate deep neural networks representations, which in theory have power expected to provide much better final classification results \citep{Joulin2016fasttext}.

%=========================================================
\section{Experiments and Analysis}\label{sec:Experiments and Analysis}

We have designed and implemented experiments to analyze the effectiveness of combining word embeddings as the text representation layer with modern classification algorithms applied to the HTC problem. After choosing an appropriate dataset for experimentation, which we describe in subsection~\ref{subsec:Exp:Dataset}, we built a data flow that transformed it depending on specific needs of each approach we describe in subsection~\ref{subsec:Exp:ClassModelsDataFlows} and train classification models using those techniques. We used each model to predict the labels of tuples left aside during the training phase to evaluate its effectiveness, which we report and analyze in subsection~\ref{subsec:Exp:Results and Analysis}.

\subsection{Dataset}\label{subsec:Exp:Dataset}

Since no dataset is widely used in the HTC research, choosing an appropriate dataset to perform HTC experiments becomes a somewhat hard task. The results from LSHTC Challenge would probably had been the best benchmark for comparison. However, as the LSHTC datasets are not available in a raw text format, they are inadequate for the purpose of this specific investigation. Therefore, we have mainly considered corpora provided by Reuters (Reuters-22173 and RCV1) and PubMed (from BioASQ). Although the PubMed collections have the advantage of containing a huge number of documents, they are ratter specialized for the medical area. We consider this as a downside for two reasons: (1) the results obtained within such a specific corpus might not generalize to other HTC tasks and (2) GloVe and word2vec pre-trained word vectors are general, which makes them inadequate for such a specific classification task\footnote{BioASQ has recently provided word embeddings pre-trained using word2vec, which could be potentially useful for this analysis. Nevertheless, as we intend to compare GloVe and word2vec results, having word embeddings trained with a single method only is not enough for our purposes.}. The Reuters collections have the advantages of including broader areas of knowledge---politics, economy, etc.---and the RCV1 \citep{lewis2004rcv1} in particular has a reasonable size with regards to number of documents (around 800K) and categories (103). RCV1 is conveniently available as a collection of XML files and publicly accessible on request for research purposes\footnote{\url{http://trec.nist.gov/data/reuters/reuters.html}}. Based on these pros and cons, we decided to use RCV1 as the experimental dataset, which contains one article per file with contents similar to example depicted in figure~\ref{fig:RCV1 example}. The example shows that the document labels are identified within XML tag \texttt{<codes class="bip:topics:1.0">}.

\begin{figure}
\centering
\fbox{\begin{minipage}{.8\linewidth}
\texttt{<?xml version="1.0" encoding="iso-8859-1" ?>\\
...\\
<headline>Tylan stock jumps; weighs sale of company.</headline>\\
...\\
<text><p>The stock of Tylan General Inc. jumped Tuesday after the maker of
process-management equipment said it is exploring the sale of the
company and added that it has already received some inquiries from
potential buyers.</p>(...)</text>\\
...\\
<metadata>\\
...\\
<codes class="bip:topics:1.0">\\
 \hspace*{1em}<code code="C15"> </code>\\
 \hspace*{1em}<code code="C152"> </code>\\
 \hspace*{1em}<code code="C18"> </code>\\
 \hspace*{1em}<code code="C181"> </code>\\
 \hspace*{1em}<code code="CCAT"> </code>\\
</codes>\\
...\\
</metadata>}
\end{minipage}}
\caption{An excerpt from a random XML file of the RCV1 dataset. \citep{lewis2004rcv1}}
\label{fig:RCV1 example}
\end{figure}

The data preparation consisted in a few steps to adequate the dataset to the machine learning algorithms. First of all, we converted those XML files into text format to remove the hypertext tags. Since we are particularly interested in the (single-label) multi-class problem, but most tuples had many labels (usually parent categories, as one can see in the example in figure \ref{fig:RCV1 example}), we counted the number of tuples per category and kept only the least frequent category of each document. This approach is based on the assumption that the least common label is the one that more specifically identifies the document. We also performed some basic homogenization, such as lower case conversion and punctuation marks removal. 

\begin{figure}
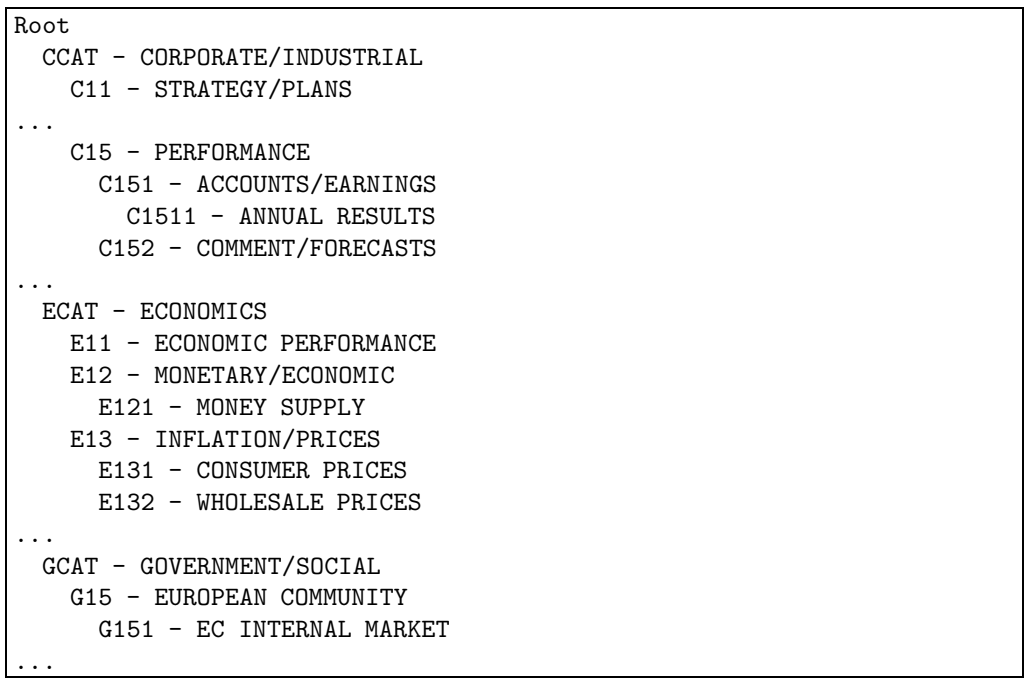

\centering
\fbox{\begin{minipage}{.8\linewidth}
\texttt{Root\\
\hspace*{1em}CCAT - CORPORATE/INDUSTRIAL\\
\hspace*{2em}C11 - STRATEGY/PLANS\\
...\\
\hspace*{2em}C15 - PERFORMANCE\\
\hspace*{3em}C151 - ACCOUNTS/EARNINGS\\
\hspace*{4em}C1511 - ANNUAL RESULTS\\
\hspace*{3em}C152 - COMMENT/FORECASTS\\
...\\
\hspace*{1em}ECAT - ECONOMICS\\
\hspace*{2em}E11 - ECONOMIC PERFORMANCE\\
\hspace*{2em}E12 - MONETARY/ECONOMIC\\
\hspace*{3em}E121 - MONEY SUPPLY\\
\hspace*{2em}E13 - INFLATION/PRICES\\
\hspace*{3em}E131 - CONSUMER PRICES\\
\hspace*{3em}E132 - WHOLESALE PRICES\\
...\\
\hspace*{1em}GCAT - GOVERNMENT/SOCIAL\\
\hspace*{2em}G15 - EUROPEAN COMMUNITY\\
\hspace*{3em}G151 - EC INTERNAL MARKET\\
...}
\end{minipage}}
\caption{An excerpt from the RCV1 topics hierarchy. \citep{lewis2004rcv1}}
\label{fig:RCV1hier}
\end{figure}

The RCV1 hierarchy consists of 104 nodes (including root) distributed among 4 levels, with 22 of them having at least one child. The target classes are distributed among all levels except for the root, all nodes are potential target classes, which indicates this corresponds to an NMLNP task. In order to compare a flat classification against hierarchical LCPN approach, we have created two data groups: (1) a train/test split by applying a holdout method \citep{han2011datamining} to randomly reserve 10\% of all RCV1 tuples for test purposes and (2) a so-called ``hierarchical split'' by recursively stratifying subsets of the train subset based on the parent nodes. At this step, all descendant tuples of a parent node were considered as part of the subset and were tagged with the label that matches the corresponding child of the node. Initial experiments with these datasets indicated the already expected incorrect classifications that occur with NMLNP in deeper hierarchy levels due to the models' inability to stop the classification before reaching a leaf node or recovering from it \citep{silla2011survey}. As a workaround, we re-executed the subset stratification by including a so-called virtual category (VC) to use the tuples from the node itself into the subset (except for the root node), as described by \citep{ying2011novel}. The final result of this hierarchical split is 22 training datasets that contained between the entire dataset (root node) and less than 0.3\% of it (node E14). The resulting datasets had a wide class imbalance variety, ranging from about 1:1 (node E51) to approximately 6000:1 (root node). Figure~\ref{fig:RCV1hier} shows an excerpt of the RCV1 topics hierarchy.

Besides using RCV1 and its hierarchy as the main elements for experimentation, we also employed general-purpose pre-trained word embeddings. The group responsible for word2vec published a dataset with around 3 million word vectors with 300 elements in length that were trained on about 100 billion words read from Google News dataset\footnote{https://code.google.com/archive/p/word2vec/}. The authors of GloVe also published pre-trained versions of word vectors; for these experiments, we will use a table with 2.2 million word vectors with 300 elements obtained from 840 billion words collected via Common Crawl\footnote{https://nlp.stanford.edu/projects/glove/}. 

\subsection{Classification Models and Data Flows for Experimentation}\label{subsec:Exp:ClassModelsDataFlows}

Out of the many possible classification models mentioned in section \ref{sec:Related Works}, we concentrated our efforts on three of them, namely fastText, XGBoost, and CNN. We have also conducted experiments using SVM to use it as a baseline for the analysis. These classifiers were trained using the two aforementioned pre-trained word embeddings as well as word embeddings obtained from the fastText supervised algorithm---more details in the upcoming paragraphs---and the following hierarchical strategies:
\begin{itemize}
\item Flat: A single model was trained as in a general multi-class task with 103 classes while ignoring the hierarchical information---see subsection~\ref{subsec:problem_criteria_solution_strategies} for more details.
\item LCPN + VC: An individual classification model was created for each one of the 22 datasets generated by the ``hierarchical split'' described in subsection~\ref{subsec:Exp:Dataset}. As a result, each model is trained with a small number of classes, including a ``virtual category'', and an extended amount of examples. For example, when learning the model for node ECAT, the local model has a few classes only (E11, E12, E13, etc. plus the ``virtual category'' ECAT). All examples from sub-nodes are be added to those classes, e.g. E13 will include all tuples from E131 and E132; the ``virtual category'' contains examples from the parent node only. During the training phase, each local model classifies learns to classify a tuple into the nodes that are immediately under it only (or back to itself in the ``virtual category''). These models are then used in a top-down strategy during the test phase. A prediction for the ``virtual category'' stops the prediction process at that level. 
%\item Flat + ancestors: the data subsets were enhanced by adding the labels for all class ancestors to each tuple, thus becoming a multi-label learning problem. For example, a tuple that belongs to class C151 will have all labels from all parent nodes, i.e. C151, C15, and CCAT. This approach was used specifically for fastText only, and takes advantage of the fact that such algorithm is able to natively handle multi-class tasks. Despite its multi-label possibilities, only the class with highest probability was considered for the evaluation.
\end{itemize}

Where computationally feasible, we generated models for both above hierarchical strategies using the same algorithm and text representation. The following list describes the learning algorithms we used with details about parameters, specific data preprocessing, and variations:

\begin{itemize}

\item fastText: We used this algorithm in two ways---as a classification learner and as a word embedding generator. As fastText is able to handle the raw text directly, no further pre-processing after the basic homogenization described in section \ref{subsec:Exp:Dataset} was necessary. We explored word embeddings with 5 different vector sizes to investigate how expanding the numerical distribution affects the final classification effectiveness in a flat strategy. During that step, we learned that fastText was able to improve its classification effectives as we increased the vector size up to a certain point only. More precisely, the \textsc{lca}F\textsubscript{1} values we found during this exploratory phase were 0.826, 0.860, 0.870, 0.871, and 0.871 for vector sizes 5, 10, 20, 30, and 40, respectively. Based on that, we decided to use the 30-element word embeddings generated during that supervised learning with other classifiers to compare this text representation against the pre-trained ones we had at hand. Other training parameters: softmax loss function, bigrams, learning rate 0.1, and 5 epochs. 

\item XGBoost: In order to accommodate the distributed text representation in a format suitable to this algorithm, we decided to combine word embeddings to compose a document representation from the corresponding word embeddings average. In other words, we took a column-wise mean of the word embeddings that corresponded to the words from each document to compound a distributed document representation. We then used the resulting average vectors as input attributes to train the classifier. This approach is similar to the one proposed by \citet{balikas2016empirical}, but considers the average only. Besides these architectures that use word embeddings, we implemented one with TF-IDF representation, created from an all-lowercase, stemmed, puntuation- and stopword-free version of the RVC1 dataset. This has the purpose of comparing the traditional TF-IDF representation directly against the distributed ones. For all experiments, we set the XGBoost algorithm to use a softmax objective and run it for 30 rounds, which seemed to be enough to converge to minimum loss. Other training parameters: learning rate 0.3 and maximum tree depth 6.

\item CNN: Since this neural network specialization has a fixed input layer size, we had to pad the corpus documents to keep only a fraction of the input text, in a similar way as described by \citet{kim2014convolutional}. An initial analysis on the corpus characteristics\footnote{Our pre-processed RCV1 training subset had an average document length of 261.57 words with 90 as the mode. Approximately 6\% of the corpus has more than 600 words only.} indicated that keeping the last 600 words would have a minimal, tolerable effect on the final classification results. We then used the Keras API\footnote{\url{https://keras.io/}}
%\citep{chollet2015keras}
to built a neural network with the following architecture: a frozen embedding layer that uses the fastText vectors, two convolution layers with rectified linear units (ReLU with kernel size 5 and 128 output filters) and max pooling (pool size 5) each \citep{Goodfellow-et-al-2016}, a densely-connected layer with ReLU activation and finally another densely-connected layer with softmax function. Other training parameters: 10 epochs, batch size 128 \citep{lai2015recurrent}.

\item SVM: We used the same document representation resulting from the word embeddings combination created for XGBoost as input attributes for an SVM classifier \cite{joachims1998text}. We used the implementation provided by package e1071\footnote{\url{https://CRAN.R-project.org/package=e1071}}
%\cite{e1071}
and all default parameters.

\end{itemize}

\subsection{Results and Analysis}\label{subsec:Exp:Results and Analysis}

During the dataset preparation, we employed a holdout method \citep{han2011datamining} to reserve a random 10\% data subset for test purposes, which we used to evaluate the models effectiveness according with the methods described in section \ref{subsec:LitRev:Evaluation}. Besides the traditional flat classification measures---precision, recall, and F\textsubscript{1}---we used their hierarchical and LCA versions to assess the models' effectiveness. We consider that \citet{kosmopoulos2015evaluation} have shown consistent results to support that \textsc{lca}F\textsubscript{1} is the most appropriate measure for HTC evaluation. We used software\footnote{HEMkit is a software tool that performs the calculation of a collection of hierarchical evaluation measures.} provided by the BioASQ team to calculate the hierarchical and LCA metrics, which are shown in table~\ref{table:Results}. 

\begin{table}[ht]
\centering
\resizebox{\textwidth}{!}{%
\begin{tabular}{ccccrrrrrrrrr}
\hline
\multirow{2}{*}{Classifier} & \multicolumn{2}{c}{Text Representation} & \multirow{2}{*}{\begin{tabular}[c]{@{}c@{}}Hierarchy\\ strategy\end{tabular}} &  \multicolumn{3}{c}{Flat (macro averaged)} & \multicolumn{3}{c}{Hierarchical} & \multicolumn{3}{c}{LCA} \\
 & Type & Size & & \multicolumn{1}{c}{P} & \multicolumn{1}{c}{R} & \multicolumn{1}{c}{F\textsubscript{1}} & \multicolumn{1}{c}{P} & \multicolumn{1}{c}{R} & \multicolumn{1}{c}{F\textsubscript{1}} & \multicolumn{1}{c}{P} & \multicolumn{1}{c}{R} & \multicolumn{1}{c}{F\textsubscript{1}} \\
\hline
CNN&sup-fastText&30&flat&0.483&0.359&0.412&0.793&0.790&0.787&0.690&0.692&0.684\\
fastText&sup-fastText&30&flat&0.700&0.491&0.577&0.900&0.901&0.899&0.873&0.874&0.871\\
fastText&sup-fastText&30&LCPN + VC&0.713&0.508&0.593&0.920&0.922&\textbf{0.920}&0.894&0.896&\textbf{0.893}\\
SVM&GloVe&300&flat&0.665&0.453&0.539&0.842&0.840&0.840&0.822&0.820&0.819\\
SVM&GloVe&300&LCPN + VC&0.679&0.534&0.591&0.857&0.856&0.855&0.839&0.838&0.836\\
SVM&sup-fastText&30&flat&0.616&0.417&0.497&0.845&0.844&0.843&0.825&0.824&0.822\\
SVM&sup-fastText&30&LCPN + VC&0.619&0.464&0.530&0.851&0.851&0.849&0.832&0.832&0.830\\
SVM&word2vec&300&flat&0.677&0.466&0.552&0.847&0.846&0.845&0.829&0.827&0.826\\
SVM&word2vec&300&LCPN + VC&0.682&0.535&0.600&0.862&0.861&0.860&0.845&0.844&0.842\\
XGBoost&GloVe&300&flat&0.182&0.720&0.290&0.830&0.822&0.824&0.787&0.768&0.771\\
XGBoost&GloVe&300&LCPN + VC&0.769&0.647&0.703&0.887&0.889&0.887&0.864&0.864&0.862\\
XGBoost&sup-fastText&30&flat&0.485&0.401&0.439&0.842&0.842&0.840&0.801&0.801&0.798\\
XGBoost&sup-fastText&30&LCPN + VC&0.706&0.558&0.623&0.874&0.878&0.875&0.846&0.850&0.846\\
XGBoost&word2vec&300&flat&0.189&0.852&0.310&0.835&0.826&0.828&0.792&0.774&0.777\\
XGBoost&word2vec&300&LCPN + VC&0.777&0.664&\textbf{0.716}&0.894&0.895&0.894&0.873&0.873&0.870\\
XGBoost&TF-IDF&~340k&flat&0.581&0.530&0.555&0.892&0.891&0.884&0.833&0.833&0.824\\
\hline
\end{tabular}%
}
\caption{Performance, Recall and F\textsubscript{1} measures in flat, hierarchical and LCA versions per classifier, text representation, and hierarchical strategy. Maximum F\textsubscript{1} values are marked in boldface.}
\label{table:Results}
\end{table}

First and foremost, the results show a considerable difference between flat and hierarchical measures. When comparing the flat F\textsubscript{1} against \textsc{lca}F\textsubscript{1}, the former has an average of 0.533, while the latter, 0.823. Moreover, despite a somewhat high Pearson correlation of 0.756 between flat F\textsubscript{1} and \textsc{lca}F\textsubscript{1}, the former is potentially misleading to assess the model effectiveness---for example, it indicates XGBoost with word2vec and LCPN+VC as the most effective model (flat F\textsubscript{1} 0.716), while it is only third best when considering \textsc{lca}F\textsubscript{1} (0.870). These evidences contribute to the ever growing understanding that flat measures are not adequate for the hierarchical context, as they insinuate a classification effectiveness well below the actual results. We also notice a high correlation between hierarchical F\textsubscript{1} and \textsc{lca}F\textsubscript{1}, with a Pearson coefficient of 0.923. We infer that this strong association occurs because the RCV1 hierarchy is only four levels deep while the main improvement offered by LCA measures in comparison to other hierarchical measures is that they prevent over-penalizing errors inflict nodes with many ancestors. This indicates that traditional hierarchical measures might be enough in low hierarchical level classification scenarios, at the same time that they are a simpler, computationally cheaper option.

\begin{figure}
\includegraphics[width=\textwidth]{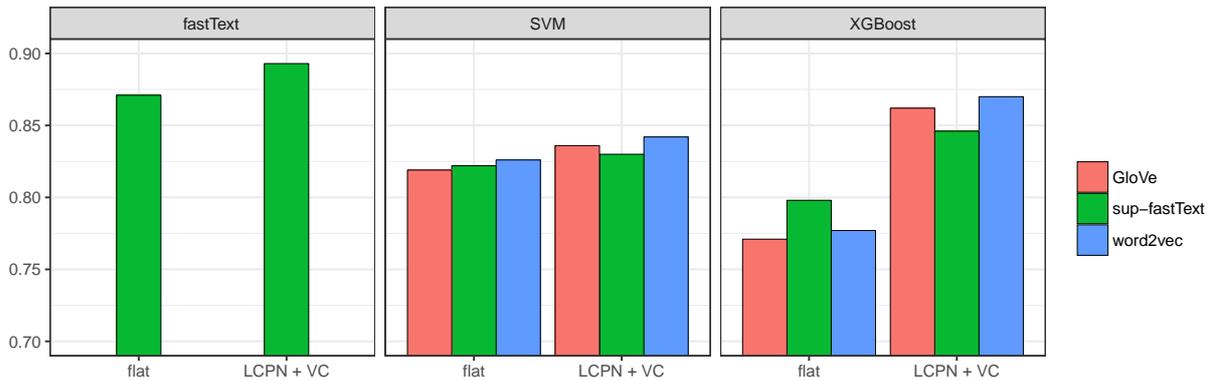}
\caption{A \textsc{lca}F\textsubscript{1} comparison per classifier, strategy, and text representation.}
\label{fig:results}
\end{figure}

The results excerpt shown in Figure~\ref{fig:results} indicates that using LCPN with VC consistently increases the \textsc{lca}F\textsubscript{1}, as all experiments using the hierarchical strategy achieve higher results then their flat counterparts. This contributes to \citet{ying2011novel} understanding that using a ``virtual category'' improves the chances to avoid misclassification by stopping the top-down approach at an adequate level. While it is clear that some classifiers take more advantage from this strategy than others---XGBoost shows a pronounced step-up---the reason for that would require further investigation. One of the hypothesis is that this hierarchical strategy helps classifiers that handle a small amount of classes better.

%This suggests that combining local classifiers with this circular strategy seems to increase the fraction of tuples placed in the right hierarchy branch over the total amount of tuples that should actually be there. However, this bouncing approach comes at a cost of decreasing the precision because some tuples get stuck at a certain level while they should be further classified at a deeper level node. On the other hand, using ancestors greatly increased the hierarchical precision---marked with a double dagger in the same table---which increases the chance that a prediction for a given hierarchy branch actually belongs to it. In summary, in both cases where we can compare flat models against their hierarchy-aware counterparts, the former are superior with regards to F\textsubscript{1}, but the latter might still suit better a particular task depending on the specific user information needs.
With regards to algorithm effectiveness, fastText is prominent among the classifiers we studied, as it exceeds all other models even when using a flat strategy. 
%It is remarkable that even 10-element vectors with very limited representational power can achieve what we consider a high effectiveness with \textsc{lca}F\textsubscript{1} greater than 0.860, which surpasses most of the other classifiers we analyzed. 
We suspect this comes from the fact that, when used in supervised mode, fastText uses the class label as learning objective, which results in word embeddings that specifically reflect the concepts behind the classes distribution. This hypothesis is supported by the fact that fastText word embeddings created in supervised mode with a relatively small amount of data yielded effectiveness on par with pre-trained word embeddings generated from much more data in an unsupervised manner. 
Considering the XGBoost algorithm, flat models using any pre-trained word embeddings are surpassed by SVM counterpart. Moreover, these XGBoost flat models had worse results than the traditional TF-IDF representation. Nevertheless, XGBoost achieved reasonable classification results and exceeded the baseline classifier in all contexts where it could take advantage of LCPN+VC strategy.
The CNN model provided a somewhat interesting result despite that we had neither thoroughly designed the architecture nor fine-tunned hyper parameters. 
%When using fastText 10-element word embeddings generated in unsupervised mode, it was considerably superior to both XGBoost models that used the same input data. Moreover, it achieved similar results using either unsupervised or supervised fastText word embeddings, which denotes that it is less dependent on the word embeddings quality. Therefore, we consider this a quite promising approach. 
Its learning phase requires much more computing resources than any of the other models we analyzed. Training this CNN--which is a rather simple implementation considering the complexity that top-notch CNN can reach---took around 6 hours in our computer (Intel\textregistered \ Core\texttrademark \ i5-4300U CPU at 1.9GHz with 8GB RAM), while typical training time for fastText and XGBoost ranged from 3 to 8 minutes for the former and 0.2 to 2.2 hours for the latter. Nevertheless, from the initial results we have found for CNN, we believe that this model can achieve a competitive level with more elaborate network configurations and given the necessary computing power.

Our results finally suggest that word embedding systems depend on the word embeddings quality to some extent, as we noticed that word2vec embeddings have a slight advantage over GloVe's. %Nevertheless, both word embeddings are clearly superior to those generated using the unsupervised fastText word embeddings. This most likely happens because we trained those unsupervised fastText word embeddings with the RCV1 train dataset only---this is a very small fraction of the amount of data that the other pre-trained word embeddings had been provided with.
At the same time, both SVM and XGBoost achieved fair results when using supervised fastText word embeddings generated from a relatively small amount of data. This contributes to the understanding word embeddings specifically generated during the classification task, even when short, are well appropriate representations for this problem.

\section{Conclusion}\label{sec:Conclusion}

Throughout this work, we have analyzed the application of distributed text representations combined with modern classification algorithms implementations to the HTC task. After an observant literature research and careful examination of related works, we identified three noticeable word embeddings generation methods---GloVe, word2vec, and fastText---and three prominent classification models---fastText, XGBoost, and CNN---that recently improved the results for the typical text classification and could potentially provide similar advancements for the hierarchical specialization. We also noticed we could exploit the hierarchical structure to build classification models using LCPN strategy and virtual categories. 

In order to assess the feasibility and effectiveness of these representations, models, and strategies to the HTC task, we performed experiments using the RCV1 dataset. By evaluating the models using flat and hierarchical measures, we confirmed that the former are inadequate for the HTC context. We also identified a strong correlation between hierarchical and LCA measures, that presumably occurs because the underlying class hierarchy of the dataset is rather shallow. The results indicate that classification models created using a hierarchical strategy (LCPN with ``virtual category'') surpasses the flat approach in all experimented equivalent comparisons. 

FastText was the outstanding method as a classifier and provided very good results as a word embedding generator, despite the relatively small amount of data provided for this second task. The algorithm seems to owe most of its superiority to the way it estimates class-oriented word embeddings in supervised mode. These findings support the increasing understanding that combining task-specific word embeddings provides the best results for text classification \citep{kim2014convolutional, tang2014learning}, to which now we include its hierarchical specialization. A direct comparison between other methods and ours is not available because previous studies have neither used hierarchical measures to assess the effectiveness or have not used the RCV1 dataset nor used a single-labeled version of it. Nevertheless, we consider the \textsc{lca}F\textsubscript{1} of 0.893 a remarkable achievement. Although some of the other classification models do not reach competitive results, they are still worth of further investigation as exploring their flexibility could still provide promising improvements.

\subsection{Future Work}\label{subsec:Exp:FutureWork}

We plan to apply these methods to the PubMed data to check how such an approach extents to the medical text context---in particular the usage of fastText. As BioASQ provides pre-trained word embeddings generated with word2vec using a considerable amount of medical texts, comparing them with those that fastText creates in supervised mode should provides us with evidence for a more general understanding on how their quality affects the final classification results. Besides that, as the Mesh hierarchy is much larger than RCV1's in all senses, it would be useful to confront the hierarchical and LCA measures in order to confirm our hypothesis about their correlation. We would also like to study the behavior of fastText when applied to task with more classes, such as the BioASQ, to check whether word embeddings with more than 30 or 40 elements would still allow for classification effectiveness improvement.
Additionally, the exploration of other text representation extensions of word2vec, like paragraph2vec and doc2vec, could complement this investigation.

Although CNN is among the models that exhibited the worst effectiveness, we believe that it deserves further investigation as this initial impression contradicts the expectations set by other studies. At the same time we recognize a deeper comprehension of its architecture is necessary to understand and apply it to the HTC context. Besides that, we would like to investigate how effective can LSTM's be with this problem, as their ability to handle sequential data matches the ordered nature of texts.

In the long term, we would like to research on the training objective used for HTC problems. We selected softmax in all the experiments, as this was the most suitable multi-class function available in the algorithm implementations we used. However, both XGBoost and CNN (through the Keras API) allow for the loss function customization. We believe that finding a differentiable function that approximates either hF\textsubscript{1} or \textsc{lca}F\textsubscript{1} and using it as the loss function rather then the softmax could finally bring together state-of-the-art algorithms with hierarchical information to create a method that implements a global HTC approach.

\section*{Acknowledgement}

This work had the support of the Brazilian National Council for the Improvement of Higher Education (CAPES) under process PROSUC 88887.150315/2017-00.

%=======================================================================
% Referências
%=======================================================================
\section*{References}
\bibliography{referencias}

\end{document}